\documentclass[letterpaper]{article} 
\usepackage{aaai2026}  
\usepackage{times}  
\usepackage{helvet}  
\usepackage{courier}  
\usepackage[hyphens]{url}  
\usepackage{graphicx} 
\usepackage{natbib}  
\usepackage{caption} 
\usepackage{algorithm}
\usepackage{algorithmic}
\usepackage{amsmath}
\usepackage{multirow}
\usepackage{amssymb}
\usepackage[table]{xcolor}
\definecolor{mygray}{gray}{0.85}
\usepackage{bbding}
\usepackage{newfloat}
\usepackage{listings}
\DeclareCaptionStyle{ruled}{labelfont=normalfont,labelsep=colon,strut=off} 
\floatstyle{ruled}
\newfloat{listing}{tb}{lst}{}
\floatname{listing}{Listing}
\title{Frequency-Driven Inverse Kernel Prediction for Single Image Defocus Deblurring}
\author{
    Ying Zhang \textsuperscript{\rm 1,2},
    Xiongxin Tang \textsuperscript{\rm 1,2}\thanks{Corresponding author.},
    Chongyi Li \textsuperscript{\rm 3},
    Qiao Chen \textsuperscript{\rm 1},
    Yuquan Wu \textsuperscript{\rm 1,2}
}
\affiliations{
    \textsuperscript{\rm 1}Institute of Software, Chinese Academy of Sciences, Beijing, China\\
    \textsuperscript{\rm 2}University of Chinese Academy of Sciences, Beijing, China\\
    \textsuperscript{\rm 3}VCIP, College of Computer
 Science, Nankai University, Tianjin, China\\

    \{zhangying2024, xiongxin, chenqiao, yuquan\}@iscas.ac.cn \\
    lichongyi@nankai.edu.cn
}

\usepackage{bibentry}

\begin{document}

\maketitle

\begin{abstract}
Single image defocus deblurring aims to recover an all-in-focus image from a defocus counterpart, where accurately modeling spatially varying blur kernels remains a key challenge.
Most existing methods rely on spatial features for kernel estimation, but their performance degrades in severely blurry regions where local high-frequency details are missing.
To address this, we propose a Frequency-Driven Inverse Kernel Prediction network (FDIKP) that incorporates frequency-domain representations to enhance structural identifiability in kernel modeling.
Given the superior discriminative capability of the frequency domain for blur modeling, we design a Dual-Branch Inverse Kernel Prediction (DIKP) strategy that improves the accuracy of kernel estimation while maintaining stability.
Moreover, considering the limited number of predicted inverse kernels, we introduce a Position Adaptive Convolution (PAC) to enhance the adaptability of the deconvolution process.
Finally, we propose a Dual-Domain Scale Recurrent Module (DSRM) to fuse deconvolution results and progressively improve deblurring quality from coarse to fine.
Extensive experiments demonstrate that our method outperforms existing approaches.
Code will be made publicly available.
\end{abstract}


\section{Introduction}
Depth of field (DOF) refers to the range around the focal plane where objects appear sharp.
Outside this range, defocus blur arises as object points form circles of confusion on the image plane.
Such blur leads to significant information loss, degrading visual quality and impairing subsequent tasks like image analysis, semantic segmentation, text recognition, and object detection. 
Therefore, single image defocus deblurring (SIDD) methods aim to restore an all-in-focus image from a defocus input.
Compared to dual-pixel image defocus deblurring (DPDD) methods \cite{DPDNet, K3DN, LDP}, which leverage pixel-wise differences between dual-pixel data to restore sharpness, SIDD is more broadly applicable but faces challenges due to the spatial variability of defocus blur kernels.

\begin{figure}
\centering{\includegraphics[width=0.47\textwidth]{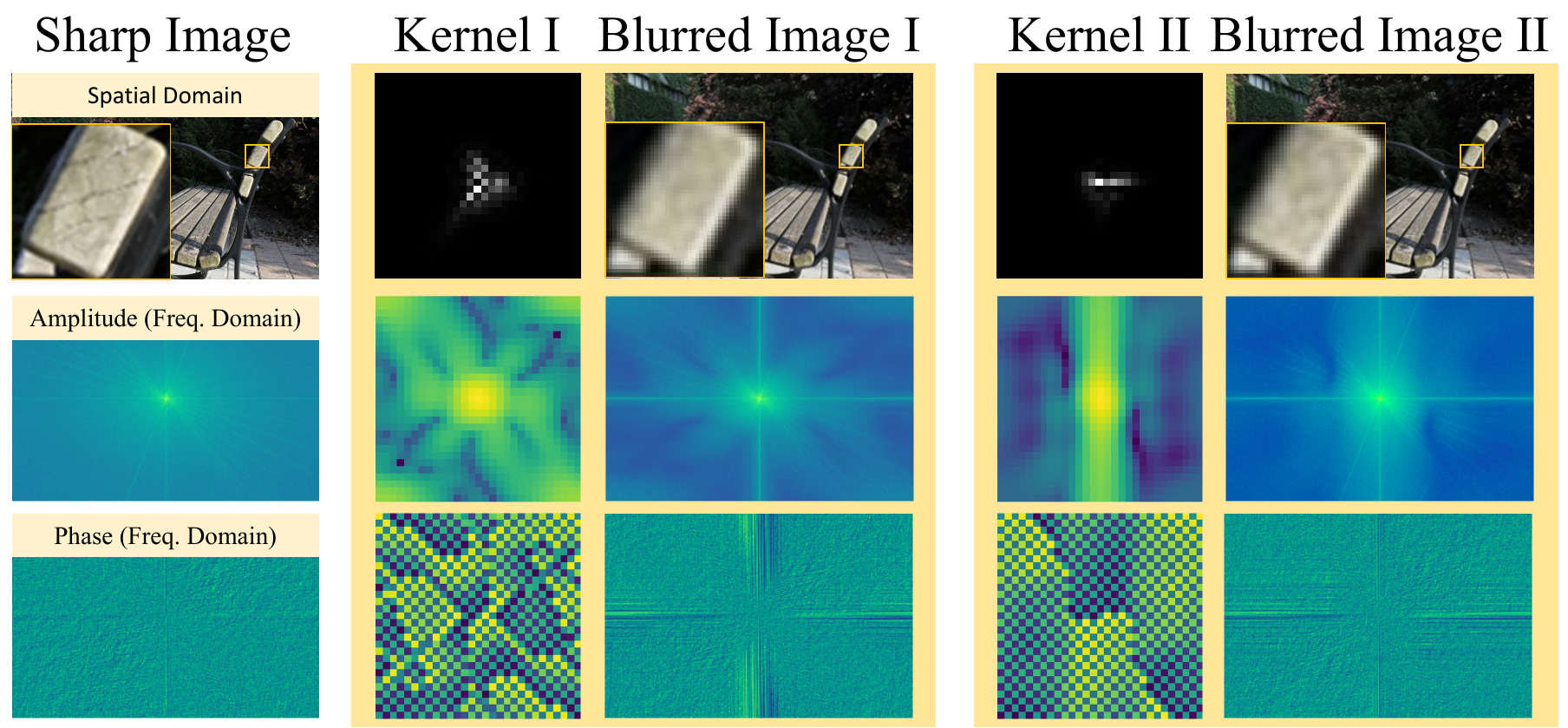}}
\caption{Compared to the visually similar blur patterns caused by different kernels in the spatial domain, frequency-domain features reveal clearer structural and discriminative properties, providing a reliable basis for kernel modeling.}
\label{fig:idea}
\end{figure}

End-to-end methods for SIDD directly learn the mapping from defocus to all-in-focus images using adaptive filtering \cite{IFAN, KPAC}, Transformer-based \cite{Restormer, Uformer, PPTFormer}, or Mamba-based models \cite{MaIR, EAMamba}. However, they heavily rely on complex DNN architectures and training data, lacking physical interpretability, which limits their robustness and generalization.
Alternatively, kernel modeling-based methods \cite{GKMNet, INIKNet, NRKNet, GGKMNet, P2IKT}  address this inverse problem from the perspective of image degradation, explicitly representing spatially varying blur kernels.
These methods attempt to improve the accuracy of blur kernel modeling through various designs, which is a key challenge in SIDD task. However, as spatial-domain methods, they rely heavily on local high-frequency information for kernel estimation, leading to reduced robustness. In regions with severe defocus blur, where local high-frequency details are suppressed, spatial features struggle to support effective kernel estimation, as shown in Fig. \ref{fig:idea}.

\begin{figure}
\centering{\includegraphics[width=0.47\textwidth]{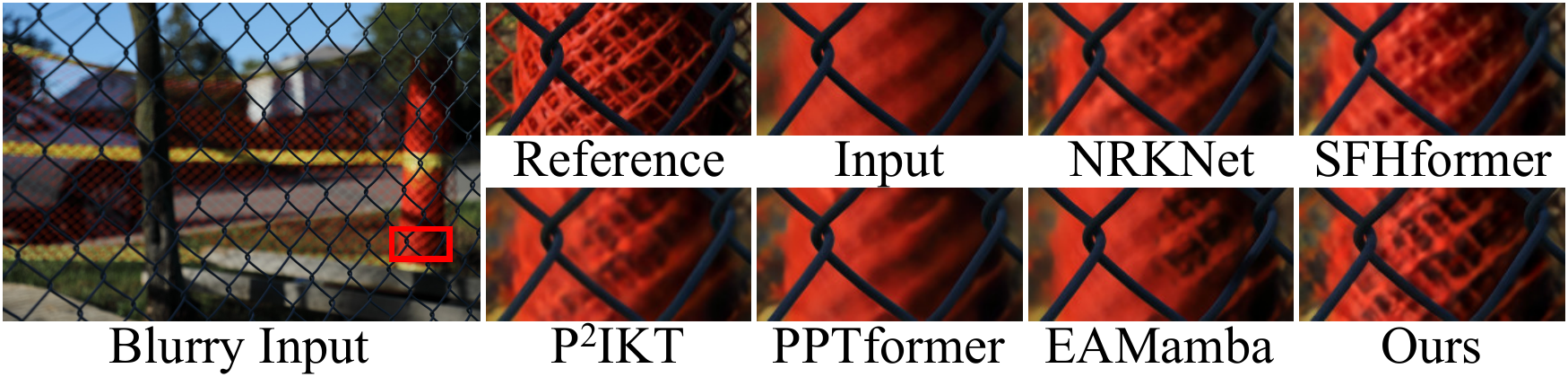}}
\caption{Visual comparison on DPDD dataset \cite{DPDNet} demonstrates that FDIKP (Ours) can effectively restore details even from severely blurry images.}
\label{fig:best_result}
\end{figure}

Previous studies have attempted to incorporate frequency information to enhance global structure modeling, such as using frequency filtering \cite{SFNet, FSNet} and frequency-aware Transformer \cite{SFHformer}. 
However, these methods typically treat frequency information as a feature enhancement tool and, like end-to-end approaches, lack physical interpretability. 
Consequently, they struggle to accurately recover structures under complex spatially varying blur kernels and severe defocus conditions.

To address the challenge, we propose a \textbf{F}requency-\textbf{D}riven \textbf{I}nverse \textbf{K}ernel \textbf{P}rediction network (\textbf{FDIKP}), comprising two main modules: the Frequency Inverse Kernel Predictor (FIKP) and the Dual-domain Scale Recurrent Module (DSRM).
Since blur can be viewed as a frequency-domain filtering process, frequency-based modeling naturally bridges degradation and restoration, offering a more discriminative representation than spatial features.
Specifically, the amplitude spectrum reflects global blur patterns, while the phase captures local details, as shown in Fig. \ref{fig:idea}.
Based on this, FIKP employs a Dual-branch Inverse Kernel Prediction (DIKP) strategy to accurately and robustly learn inverse kernels from the amplitude-phase representation.
Given the limited number of predicted kernels, we introduce a Position Adaptive Convolution (PAC), which dynamically adjusts the receptive field of inverse kernels based on prior knowledge \cite{KPAC} that defocus kernels vary mainly in scale but share similar shapes, enabling pixel-wise adaptive deconvolution.
We further propose the DSRM, which predicts coefficient maps to aggregate the results from FIKP. DSRM effectively leverages the complementary strengths of spatial-domain operations in capturing local features and frequency-domain operations in modeling global structures, thereby enhancing the representation from local to global, ultimately leading to high-quality aggregation.
Finally, deblurring process is performed in a coarse-to-fine manner within a scale recurrent architecture, yielding high-quality deblurring results.
Extensive experiments demonstrate that our method achieves competitive performance, effectively restoring structure and details in severely blurry defocus images, as shown in Fig. \ref{fig:best_result}.

The main contributions of this paper are as follows:
\begin{itemize}
    \item We propose FDIKP, a frequency-driven framework that bridges degradation and restoration, enabling more effective handling of spatially varying defocus blur.
    \item We propose FIKP, a novel frequency-domain inverse kernel prediction model that exploits frequency-domain characteristics for accurate and stable blur modeling, while integrating prior knowledge to establish pixel-level kernel associations.
    \item We propose DSRM to achieve effective dual-domain feature collaboration and progressive multi-scale refinement, thereby enhancing deblurring performance.
    \item Extensive experiments demonstrate that the proposed method achieves superior performance across multiple evaluation metrics and datasets.
\end{itemize}

\section{Related Work}
\subsection{Single Image Defocus Deblurring}
Traditional two-stage methods \cite{EBED, JNB, DMENet} for SIDD first estimate a defocus map to capture the spatially varying blur, followed by non-blind deconvolution to recover the sharp image. 
However, even slight errors in DM estimation can propagate and degrade deblurring performance.

End-to-end deep learning approaches address spatially varying defocus using adaptive filters \cite{IFAN} or atrous convolutions \cite{KPAC}.
Recently, Transformer-based \cite{Restormer, PPTFormer} and Mamba-based \cite{MaIR, EAMamba} models enhance global context modeling to better restore image details.
However, these methods rely heavily on training data and suffer from limited robustness.

Kernel modeling-based methods explicitly model the blur kernel and apply inverse techniques for deblurring.
GKMNet \cite{GKMNet} and GGKMNet \cite{GGKMNet} represent defocus kernels using a Gaussian kernel mixture model.
To achieve more general kernel representations, NRKNet \cite{NRKNet} proposes a learnable recursive kernel representation, while other methods leverage implicit neural representation \cite{INIKNet} and kernel prediction network \cite{P2IKT} to estimate inverse kernels.
Unlike these methods, which improve the accuracy of the blur kernel representation only in the spatial domain, our approach leverages the properties of the blur kernel in the frequency domain to achieve robust and accurate kernel prediction. 

\begin{figure*}
\centering{\includegraphics[width=\textwidth]{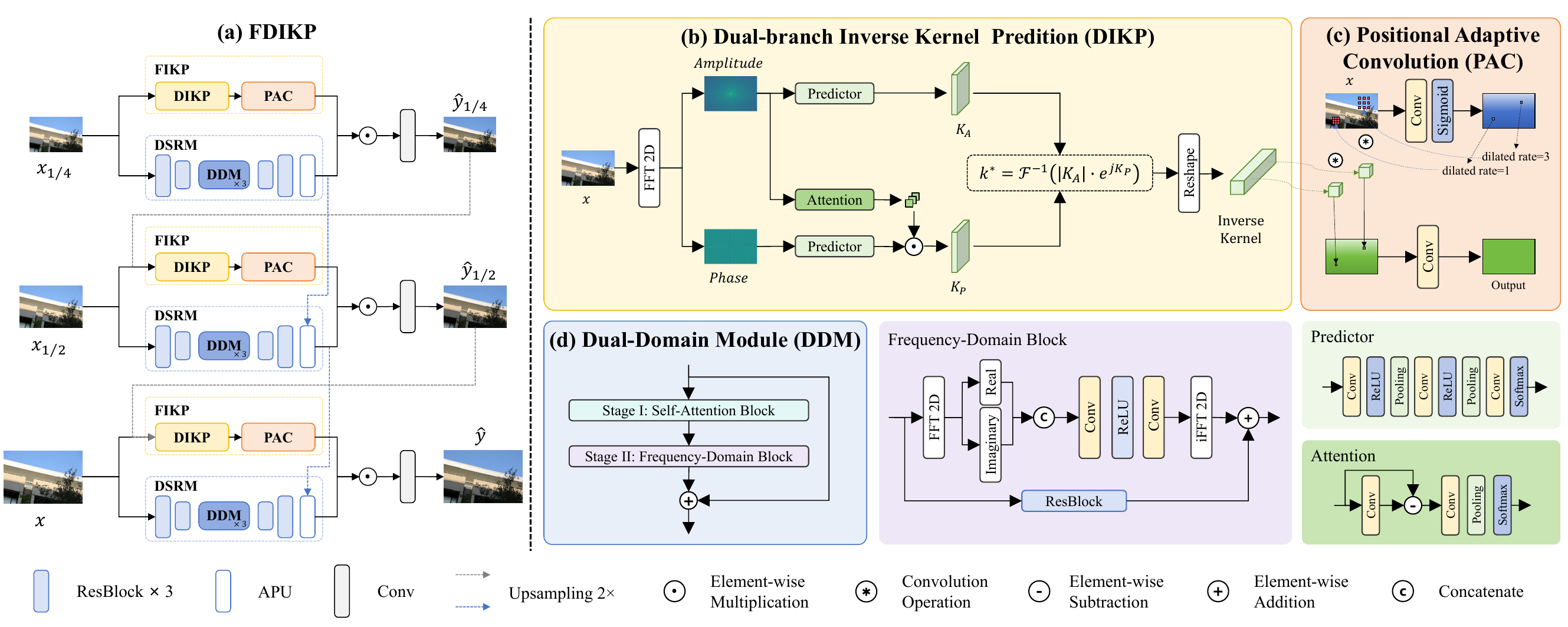}}
\caption{(a) Overall architecture of \textbf{FDIKP}, a three-stage scale recurrent network with FIKP (DIKP+PAC) and DSRM. (b) \textbf{DIKP} accurately models the inverse kernel by amplitude-phase dual-branch structure. (c) \textbf{PAC} adaptively adjusts the receptive field to establish a pixel-wise correspondence between the inverse kernel and the image. (d) \textbf{DDM} achieves dual-domain collaboration through a two-stage structure: the first stage extracts high-frequency details, and the second stage complements this by modeling global structures while retaining the extracted information.}
\label{fig:model}
\end{figure*}

\subsection{Frequency-Based Deblurring}
Frequency domain information has achieved remarkable results in image deblurring \cite{EFEP,SFNet,FSNet,LoFormer,SFHformer}.
EFEP \cite{EFEP}, SFNet \cite{SFNet}, and FSNet \cite{FSNet} exploited frequency decomposition to suppress irrelevant information and enhance features. 
LoFormer \cite{LoFormer} and SFHformer \cite{SFHformer} incorporate frequency priors to build frequency-aware Transformers.
However, these methods essentially utilize frequency information as an auxiliary feature enhancement.
Therefore, they remain black-box architectures with strong reliance on training data, exhibiting limited robustness and generalization under complex spatially varying kernels or severe defocus blur.
In contrast, we adopt a frequency-domain modeling framework rooted in the physical degradation process to accurately predict inverse kernels.
Compared to data-driven methods, our physics-guided approach enhances interpretability and generalization, especially in severely blurred regions with better structure preservation and fewer artifacts.

\subsection{Scale Recurrent Architecture}

Scale-recurrent architecture \cite{SRA1} has been widely adopted for image deblurring using a coarse-to-fine strategy.
SRN \cite{SRN} introduced ConvLSTM to enhance information flow across scales. 
Based on this, GKMNet \cite{GKMNet} incorporated attention mechanisms, while P$^2$IKT \cite{P2IKT} integrated Transformer to strengthen feature representation.
Built upon this backbone, our DSRM innovatively introduces a frequency-domain block to enable spatial-frequency collaboration.
Unlike prior works focusing solely on spatial features, our dual-domain design leverages the global modeling capacity of frequency information to compensate for high-frequency loss.

\section{Methodology}

\subsection{Motivation}

In image deblurring, the spatial structure of the blur kernel is difficult to observe directly from the blurry image (see the 1st row of Fig. \ref{fig:idea}), which undermines the robustness of spatial-domain kernel estimation. Since blurring is inherently a frequency-domain filtering process, we instead explore kernel modeling in the frequency domain.

Following common practice, we model the degradation process from a sharp image $y$ to a blurry image $x$ as:
\begin{equation}
    x = y \otimes k,
\label{equation_1}
\end{equation}
where $k$ denotes blur kernel and $\otimes$ represents convolution operation. 
Through fast Fourier transform (FFT) \cite{fft}, an image can be decomposed in the frequency domain into two components: the amplitude spectrum and the phase spectrum. 
As illustrated in Fig.~\ref{fig:idea}, the amplitude spectrum of a blurry image clearly encodes the structural characteristics of the blur kernel, making it easier to direct learning. 
In contrast, the phase spectrum is less intuitively interpretable, but it encodes fine-grained structural cues and local details, which play an indispensable role in accurately reconstructing the spatial-domain kernel via the inverse fast Fourier transform (iFFT) :
\begin{equation}
    k = \mathcal{F}^{-1}(K) = \mathcal{F}^{-1}\left(|K|\cdot e^{j\phi_K}\right),
    \label{eq:4}
\end{equation}
where, $K=\mathcal{F}(k)$ is the frequency-domain representation of $k$, $\mathcal{F}^{-1}(\cdot)$ denotes the iFFT operation, $|\cdot|$ and $\phi_{(\cdot)}$ denote amplitude and phase spectra, respectively.
Neglecting phase information (i.e., assuming $\phi_K=0$) introduces random phase errors during reconstruction, leading to kernel distortion.
Therefore, we leverage the more accessible amplitude spectrum as a structural prior to facilitate frequency-domain learning, particularly aiding the estimation of the phase component.

Furthermore, in the image deblurring task, our objective is to predict the restored image $\hat{y}$ by performing deconvolution of the blurry image using the inverse kernel $k^{\ast}$:
\begin{equation}
    \hat{y} = x \otimes k^{\ast},\,\text{where}\, k^{\ast} = \mathcal{F}^{-1}(\frac{1}{|K|}\cdot e^{-j\phi_K}).
    \label{eq:deconv}
\end{equation}
This indicates that the amplitude spectrum of $k^{\ast}$ equals $\frac{1}{|K|}$, while its phase spectrum corresponds to $-\phi_K$. 
Analogous to blur kernel estimation, accurate inverse kernel estimation can therefore be achieved through joint modeling of amplitude and phase spectra.

\subsection{Overall Architecture}

As illustrated in Fig. \ref{fig:model} (a), the FDIKP adopts a Scale Recurrent Architecture that progressively refines the image from coarse to fine through three recurrent stages operating at different resolutions. 
Specifically, the network processes the downsampled inputs $x_{1/4}$, $x_{1/2}$, and the original resolution $x$ in succession to reconstruct increasingly refined outputs $\hat{y}_{1/4}$, $\hat{y}_{1/2}$, and the final prediction $\hat{y}$.
Each stage consists of two main modules, FIKP and DSRM, and ensures information propagation across stages through the hidden state.

Taking the second stage as an example, the upsampled output $\hat{y}_{1/4}$ in the first stage and the stage input $x_{1/2}$ are independently processed by FIKP, and their resulting features are concatenated to form the feature maps $F_{1/2}$.
Simultaneously, $x_{1/2}$ is fed into DSRM to generate the corresponding coefficient maps $C_{1/2}$.
The output $\hat{y}_{1/2}$ at this scale is obtained by weighting $F_{1/2}$ with $C_{1/2}$, followed by a convolutional layer to adjust the channel dimensions.

We optimize the proposed network FDIKP based on the following multi-scale loss function:
\begin{equation}
    \mathcal{L} = \lambda_1\mathcal{L}(y,\hat{y})+\lambda_2\mathcal{L}(y_{1/2},\hat{y}_{1/2})+\lambda_3\mathcal{L}(y_{1/4},\hat{y}_{1/4}),
\end{equation}
where $y$ denotes the ground truth image, and $y_{1/2}$ and $y_{1/4}$ represent its 2× and 4× downsampled versions, respectively; $\lambda_1$, $\lambda_2$, and $\lambda_3$ are hyperparameters for the multi-scale loss, set to 1, 0.2, and 0.1, respectively, in the experiments.
The single-scale loss $\mathcal{L}(y,\hat{y})$ is defined as follows:
\begin{equation}
    \mathcal{L}(y,\hat{y})=\alpha \mathcal{L}_{2}(y,\hat{y})+\beta \mathcal{L}_\text{lpips}(y,\hat{y})+\gamma \mathcal{L}_\text{freq}(y,\hat{y}),
\end{equation}
where $\alpha$, $\beta$, and $\gamma$ are hyperparameters that adjust the weight of each loss function, which are set to 1, 0.2, and 0.2, respectively, in the experiment. 
Specifically, the pixel-level loss ($\mathcal{L}2$) minimizes the pixel-wise difference between the target and predicted images, ensuring the recovery of fine image details. 
Meanwhile, $\mathcal{L}_\text{lpips}$ uses LPIPS \cite{lpips} as a loss function to improve the perceptual quality.
Additionally, we include frequency loss ($\mathcal{L}_\text{freq}$), which measures the difference in the frequency domain to ensure the network maintains global structures that might be overlooked by pixel-level loss.

\subsection{Frequency Inverse Kernel Predictor}

Building upon the frequency-domain inverse kernel modeling introduced earlier, we develop a dual-branch architecture that simultaneously estimates initial inverse kernels from both amplitude and phase spectra. 
Considering the high computational cost of modeling a unique inverse kernel at every spatial location, we estimate only a limited set of kernels.
To capture spatially varying blur patterns, inspired by the observation \cite{KPAC} that defocus kernels exhibit consistent shapes but varying scales, we design a position-adaptive mechanism.
This mechanism dynamically adjusts the scale of inverse kernels, enabling flexible kernel-adaptive restoration using limited kernels.
The resulting FIKP integrates dual-branch inverse kernel prediction (DIKP) with position adaptive convolution (PAC), as illustrated in Fig. \ref{fig:model} (b)(c).

\subsubsection{Dual-branch Inverse Kernel Prediction}
Given input image $x$, we compute its frequency components, the amplitude spectrum $A$ and phase spectrum $P$ via FFT.
In the amplitude branch, we employ a Predictor to estimate the amplitude-guided kernel $K_A$:
\begin{equation}
    K_A = \text{Predictor}(A),
\end{equation}
where Predictor consists of two structural units, each comprising a Conv, ReLU, and Adaptive Average Pooling, followed by a final Conv and Softmax activation.
In the phase branch, however, the kernel structure is less distinct, making the estimation more challenging. 
To address this, we first obtain an initial estimate using the Predictor. 
Then, guided by the amplitude spectrum, we apply a spatial Attention mechanism to modulate the initial estimate, yielding the phase-guided kernel $K_P$:
\begin{equation}
    K_P = \text{Predictor}(P) * \text{Attention}(A),
\end{equation}
where Attention first computes the amplitude residual to filter out kernel-irrelevant information. It then applies Conv, Global Average Pooling, and Softmax activation to obtain the final attention map.
The final inverse kernel $k^{\ast}$ is reconstructed as follows:
\begin{equation}
    k^{\ast} = \mathcal{F}^{-1}(|K_A|.e^{jK_P'}).
\end{equation}

\subsubsection{Position Adaptive Convolution}
To generate accurate deconvolution features with a limited set of inverse kernels, we design a position adaptive mechanism inspired by dilated convolution \cite{DC, SDWNet, FADC}.
Compared to standard dilated convolution that lacks semantic awareness of blur scale, our method adaptively adjusts the receptive field of the predicted inverse kernels based on the learned defocus level, enabling more precise deconvolution.
Specifically, we extract a blur-aware dilated map $D$ from the input image through Conv and Sigmoid, capturing the spatially varying defocus.
This map adaptively configures the dilated rate for applying the inverse kernel during deconvolution:
\begin{equation}
    F(p)=\sum_{i=1}^{K\times K} k^{\ast}_i x(p+\Delta p_i\times D(p)),
\end{equation}
where $F(p)$ represents the pixel value in the preliminary deconvolution feature maps, $K$ is the size of the inverse kernel, $k^{\ast}_i$ represents the kernel weights, $\Delta p_i$ represents the $i$-th location of the pre-defined grid sampling (-1,-1), (-1, 0), ..., (+1,+1), and $D(p)$ is the dilated rate at $p$.
To mitigate estimation errors, we refine the preliminary deconvolution features through an additional Conv layer, yielding the final deconvolution feature maps.

\begin{table*}
    \centering
    \begin{tabular}{c|c c c | c c c | c c c | c c }
    \hline
        \multirow{2}{*}{\textbf{Method}} & \multicolumn{3}{c|}{\textbf{DPDD}} & \multicolumn{3}{c|}{\textbf{RealDOF}} & \multicolumn{3}{c|}{\textbf{RTF}} & \# Params  & Time \\
		& PSNR & SSIM & LPIPS & PSNR & SSIM & LPIPS & PSNR & SSIM & LPIPS & (MB) & (s)\\ \hline
        Blurry Input & 23.89 & 0.725 & 0.349 & 22.33 & 0.633 & 0.524 & 24.18 & 0.739 & 0.339 & - & - \\
        DPDNet (2020) & 23.91 & 0.730 & 0.357 & 22.52 & 0.644 & 0.583 & 22.95 & 0.754 & 0.388 & 35.25 & 0.31 \\
        IFAN (2021) & 25.36 & 0.788 & 0.216 & 24.71 & 0.748 & 0.304 & 24.92 & 0.821 & 0.214 & 10.18 & 0.20\\
        DRBNet (2022) & 25.36 & 0.787 & 0.239 & 24.34 & 0.743 & 0.335 & 24.56 & 0.808 & 0.261 & 11.69 & \textbf{0.01}\\
        NRKNet (2023) & 26.11 & 0.803 & 0.223 & 25.03 & 0.752 & 0.335 & \underline{25.92} & 0.846 & 0.204 & 6.09 & 0.42 \\
        SFHformer (2024) & 26.14 & \underline{0.808} & 0.220 & 25.18 & \underline{0.761} & 0.326 & 25.59 & 0.844 & 0.227 & 5.83 & 0.42 \\
        P$^2$IKT (2024) & \underline{26.24} & 0.805 & 0.194 & \underline{25.43} & 0.758 & 0.307 & 25.83 & 0.841 & 0.201 & \textbf{3.32} & \underline{0.05} \\
        PPTformer (2025) & 26.09 & 0.805 & 0.192 & 25.27 & \textbf{0.771} & \underline{0.299} & 25.42 & \underline{0.850} & \underline{0.182} & 39.65 & 2.78  \\
        EAMamba (2025) & 26.02 & \textbf{0.813} & \textbf{0.181} & 24.59 & \underline{0.761} & 0.309 & 24.36 & 0.817 & 0.231 & 25.32 & 2.76  \\
        \rowcolor{mygray}
        \textbf{Ours} & \textbf{26.42} & \textbf{0.813} & \underline{0.185} & \textbf{25.58} & \textbf{0.771} & \textbf{0.283} & \textbf{26.52} & \textbf{0.865} & \textbf{0.163} & \underline{5.52} & 0.10 \\
        \hline
    \end{tabular}
    \caption{Quantitative comparison results on \textbf{DPDD} \cite{DPDNet}, \textbf{RealDOF} \cite{IFAN} and \textbf{RTF} \cite{RTF} dataset. \textbf{Bold} and \underline{underline} indicate the best and second-best results, respectively.}
\label{tab:results}
\end{table*}

\subsection{Dual-Domain Scale Recurrent Module}

The DSRM is designed to predict the coefficient maps used to fuse the deconvolution feature maps from FIKP. 
As illustrated in Fig. \ref{fig:model}(a), DSRM follows an encoder-decoder architecture. 
In the encoder, multi-scale features are extracted using ResBlocks \cite{ResNet}.
These features are then passed through a bottleneck, where a Dual-Domain Module (DDM) is introduced to enhance the representation.
These enhanced features are passed to the decoder and combined with the corresponding encoder features for further processing. 
Finally, the decoded features are fed into a recurrent unit, APU \cite{APU, GKMNet}, which integrates them with the hidden state from the previous stage to estimate the coefficient maps and update the hidden state for the next stage.

In DSRM, the capability to effectively model global and local features is vital for achieving high prediction accuracy. 
Given the inherent global modeling capability of frequency information, we incorporate it to improve representation quality and design a two-stage Dual-Domain Module (DDM). 
Unlike previous spatial-frequency dual-domain designs that adopt a parallel structure to separately enhance spatial and frequency features before fusion, we observe that a sequential modeling strategy yields better results. 
Therefore, the proposed DDM is structured in two stages: the first stage employs a spatial self-attention block \cite{Stripformer} to capture high-frequency details, while the second stage leverages a frequency-domain block to extract global structures, as illustrated in Fig. \ref{fig:model}(d). 
As shown in the purple part of Fig. \ref{fig:model}, the frequency-domain block first transforms the features into the frequency domain via FFT, enhances them with a Conv-ReLU-Conv block, and then transforms them back to the spatial domain via iFFT. A conventional ResBlock is also included to preserve the learned local high-frequency components. To balance computational cost, DDM is applied only at the bottleneck.

\section{Experiments}

\subsection{Experimental Setting}
We train and evaluate our model on the DPDD dataset \cite{DPDNet}, which contains 500 scenes captured using a DSLR camera. 
To further assess the generalization ability of our model, we conduct additional evaluations on three benchmark datasets: RealDOF \cite{IFAN}, RTF \cite{RTF}, and CUHK \cite{CUHK}. 
RealDOF is another paired dataset captured with a DSLR camera, containing 50 scenes. 
RTF is a synthetic paired dataset generated using a light field camera, consisting of 22 scenes. 
CUHK comprises 704 real-world images collected from the internet, covering diverse scenes, but does not provide paired ground-truth images.

Our model is implemented using PyTorch and trained on an NVIDIA GeForce RTX 3090 GPU.
Unless otherwise specified, the model predicts 5 inverse kernels of fixed size 5$\times$5 in the experiment.
Our network is trained using a two-stage strategy for a total of 1500 epochs with the Adam optimizer ($\beta_1=0.9,\beta_2=0.999$) \cite{adam}.
The initial learning rate is set to $1\times 10^{-4}$ and dynamically adjusted using the MultiStepLR scheduler.
To enhance the diversity of the training data, we apply data augmentation techniques, including random cropping, horizontal and vertical flipping, and 90° rotation.

Multiple metrics are used to evaluate the performance of the SIDD task, including Peak Signal-to-Noise Ratio (PSNR) \cite{psnr}, Structural Similarity Index Measure (SSIM) \cite{ssim}, and Learned Perceptual Image Patch Similarity (LPIPS) \cite{lpips}. Additionally, we report the number of model parameters and inference time to provide a more comprehensive evaluation. Inference time is measured as the average over DPDD test set.

\renewcommand{\dblfloatpagefraction}{.9}
\begin{figure*}
\centering{\includegraphics[width=\textwidth]{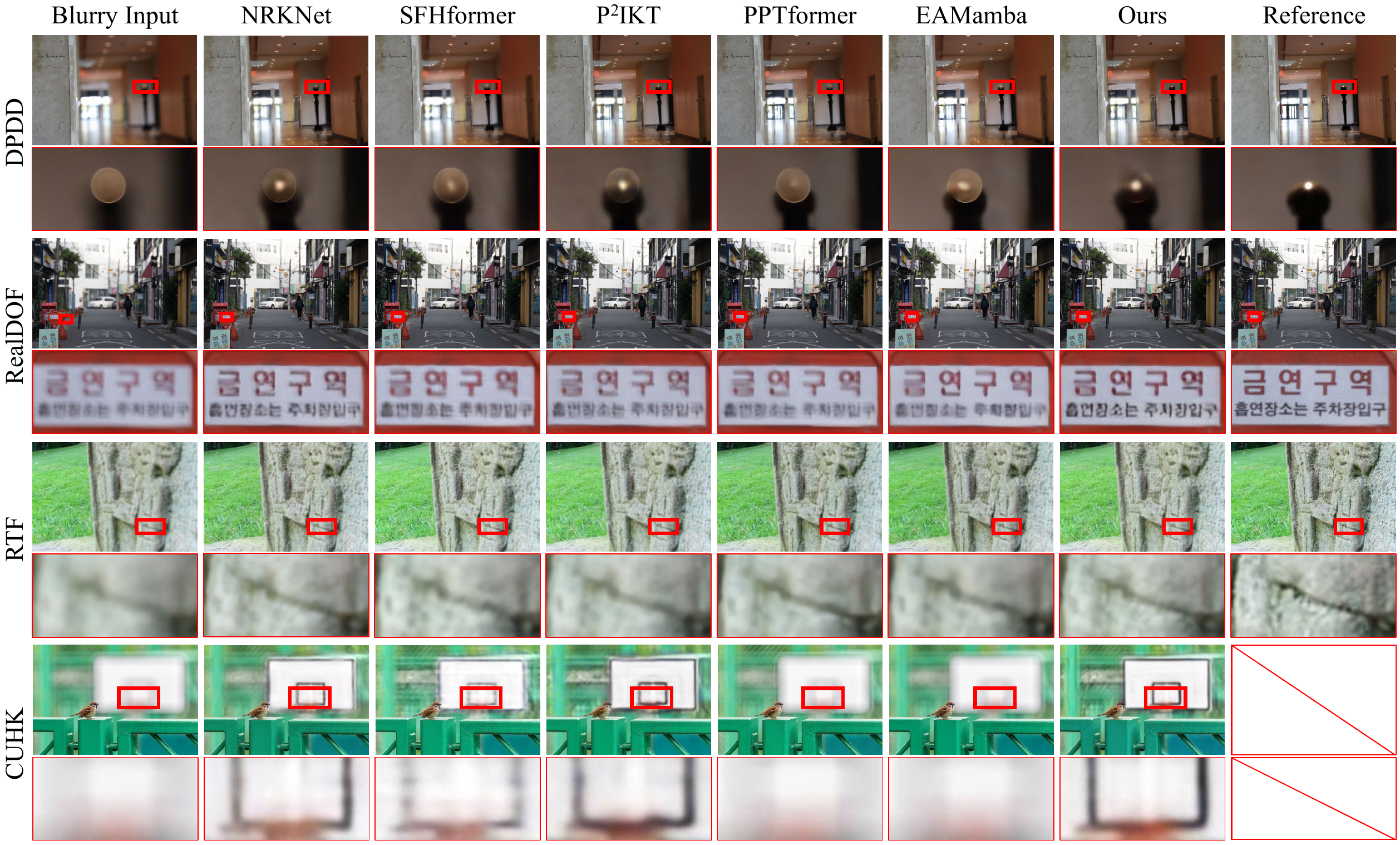}}
\caption{Qualitative results on \textbf{DPDD} \cite{DPDNet}, \textbf{RealDOF} \cite{IFAN}, \textbf{RTF} \cite{RTF}, and \textbf{CUHK} \cite{CUHK} datasets, including \textbf{NRKNet} \cite{NRKNet}, \textbf{SFHformer} \cite{SFHformer}, \textbf{P$^2$IKT} \cite{P2IKT}, \textbf{PPTformer} \cite{PPTFormer}, \textbf{EAMamba} \cite{EAMamba}, and \textbf{our method}.}
\label{fig:results}
\end{figure*}

\subsection{Deblurring Performance Evaluation}
To demonstrate the superiority of our model, we compare our FDIKP with currently advanced methods, including DPDNet \cite{DPDNet}, IFAN \cite{IFAN}, DRBNet \cite{DRBNet}, NRKNet \cite{NRKNet}, SFHformer \cite{SFHformer}, P$^2$IKT \cite{P2IKT}, PPTformer \cite{PPTFormer}, and EAMamba \cite{EAMamba}.
For methods with available pretrained weights, we directly use their released models for comparison.
Since neither DPDNet nor SFHformer released pretrained models for the SIDD task, and DRBNet was originally trained on both DPDD and LFDOF \cite{LFDOF} datasets, we retrain them on the DPDD dataset for a fair comparison.

Table \ref{tab:results} shows the quantitative results on the DPDD test set. Our method achieves the highest PSNR and SSIM, reaching 26.42 dB and 0.813, respectively.
P$^2$IKT improves PSNR over the input by 2.35 dB, while our method further increases it by 0.18 dB (7.7\%), resulting in a total gain of 2.53 dB.
For SSIM, SFHformer achieves an improvement of 0.083, whereas our method further improves it to 0.088, a 6\% relative increase.
Although our method ranks second in LPIPS (0.185 vs. EAMamba's 0.181, it significantly outperforms in terms of model efficiency. Specifically, our model has only 5.52 MB parameters, which is 21.8\% of EAMamba's size, and achieves an inference time of just 0.10 seconds, representing a 96.4\% reduction in runtime. This highlights its advantage in both compactness and efficiency.
The first row of Fig. \ref{fig:results} illustrates that our method better restores sharp light sources by removing defocus halos, outperforming others in reducing aperture artifacts.
More visual results are available in the supplementary.

\subsection{Generalization Ability Analysis}
To evaluate the generalization and robustness of our method, we conduct additional experiments on other benchmark datasets: RealDOF, RTF, and CUHK. 
Images in the RTF dataset are cropped to a size of 352$\times$352 to prevent inference errors caused by the original image dimensions.

Table \ref{tab:results} reports quantitative results on RealDOF and RTF datasets, where FDIKP consistently achieves state-of-the-art performance across all metrics.
On RealDOF dataset, the second-best results in terms of PSNR (P2IKT), SSIM (SFHformer and EAMamba), and LPIPS (PPTformer) achieve improvements over the input image by 3.1 dB, 0.128, and 0.225, respectively. Our method achieves additional gains of 4.8\%, 7.8\%, and 7.1\%, respectively.
On RTF dataset, the second-best methods NRKNet for PSNR and PPTformer for SSIM and LPIPS achieve improvements over the input image by 1.74 dB, 0.111, and 0.157, respectively. Our model further improves upon these by 34.5\%, 13.5\%, and 12.1\%, respectively.
The second and third rows of Fig. \ref{fig:results} show the visual comparisons on RealDOF and RTF datasets, respectively. Leveraging pixel-wise spatially aligned inverse kernels, our method produces sharper images with clearer structural details, particularly around edges such as text contours and narrow gaps.
The last row of Fig. \ref{fig:results} presents the visual comparison results on CUHK dataset. Compared with other methods, our approach leverages frequency-domain modeling to enhance robustness under severe blur, leading to better restoration of fine details, such as the lines of the basketball hoop.
Overall, FDIKP exhibits strong generalization across diverse real-world conditions.
Failure cases under extremely blurry scenes are discussed in the supplementary material.

\begin{table*}
    \centering
    \tabcolsep=0.3cm
    \begin{tabular}{c c c c | c c c c | c }
    \hline
        \multicolumn{2}{c}{FIKP} & \multirow{2}{*}{DSRM} & \multirow{2}{*}{SRAM} & \multirow{2}{*}{PSNR} & \multirow{2}{*}{SSIM} & \multirow{2}{*}{MAE} & \multirow{2}{*}{LPIPS} & \multirow{2}{*}{\# Params (MB)} \\ 
        DIKP & PAC & ~ & ~ & ~ & ~ & ~ & ~  \\ \hline
        ~ & ~ & ~ & \checkmark & 25.49 & 0.784 & 0.0409 & 0.216 & 5.51 \\
        \checkmark & \checkmark & ~ & \checkmark & 25.89 & 0.797 & 0.0385 & 0.205 & 5.62 \\
        ~ & ~ & \checkmark & ~ & 26.11 & 0.803 & 0.0373 & 0.195 & 5.41 \\
        \checkmark & ~ & \checkmark & ~ & 26.28 & 0.809 & 0.0368 & 0.193 & 5.52 \\
        \rowcolor{mygray}
        \checkmark & \checkmark & \checkmark & ~ & 26.42 & 0.813 & 0.0366 & 0.185 & 5.52 \\
        \hline
    \end{tabular}
    \caption{Results of ablation study on DPDD dataset. All study is conducted under the scale recurrent framework.}
\label{tab:ablation}
\end{table*}

\subsection{Ablation Study and Analysis}
\subsubsection{Effectiveness of Each Component in FDIKP}
To validate the contribution of each component in FDIKP, i.e., FIKP (DIKP+PAC) and DSRM, we conduct an ablation study on the DPDD dataset.
The baseline model is built using SRAM \cite{GKMNet} within a scale-recurrent framework.
We double the channels in each SRAM block and add a ResBlock at the bottleneck to maintain a comparable parameter count.
Based on this, we validate the effect of FIKP and DSRM by introducing FIKP and replacing SRAM with DSRM. 
Then, starting from the scale recurrent framework with DSRM, we gradually add the DIKP and PAC. 

\begin{figure}
\centering{\includegraphics[width=0.47\textwidth]{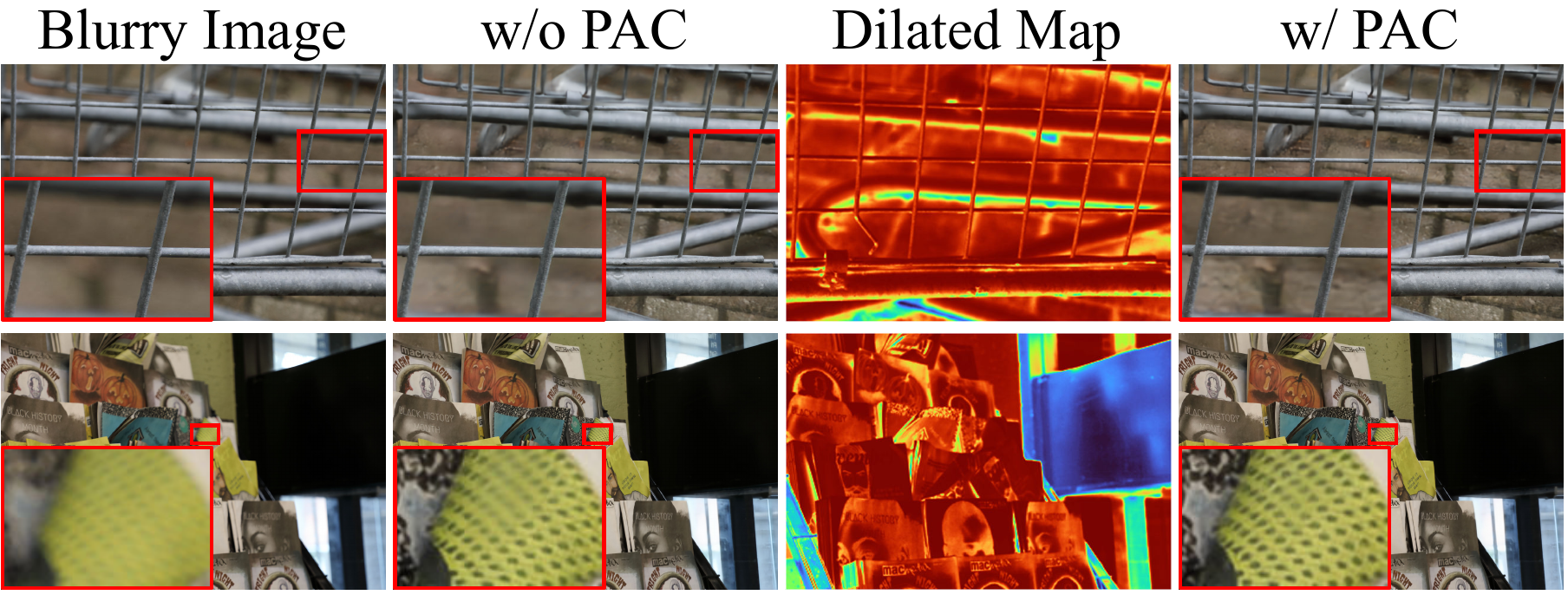}}
\caption{Deblurring results w/o and w/ PAC, with visualization of the dilated map in PAC. Redder regions indicate larger inverse kernel receptive fields, typically found in smoother areas of the blurry image.}
\label{fig:dilation_map}
\end{figure}

As shown in Table \ref{tab:ablation}, both FIKP and DSRM significantly improve performance.
FIKP increases PSNR from 25.49 to 25.89 dB, demonstrating its effectiveness in frequency-domain inverse kernel modeling.
DSRM further boosts the PSNR to 26.11 dB, indicating that the dual-domain design effectively leverages the global modeling capability of the frequency domain. 
Within FIKP, DIKP accurately predicts inverse kernels using frequency information, raising the PSNR to 26.28 dB. 
PAC introduces almost no extra parameters to overcome the limitation of a fixed number of inverse kernels, further improving the PSNR to 26.42 dB.
As shown in Fig. \ref{fig:dilation_map}, PAC perceives the degree of blur in the image and adaptively adjusts the receptive field of the inverse kernel accordingly. This enables a pixel-wise correspondence between the inverse kernel and the image, which effectively enhances deblurring performance.

\subsubsection{Analysis of Kernel Modeling in Frequency Domain}
\begin{figure}
\centering{\includegraphics[width=0.47\textwidth]{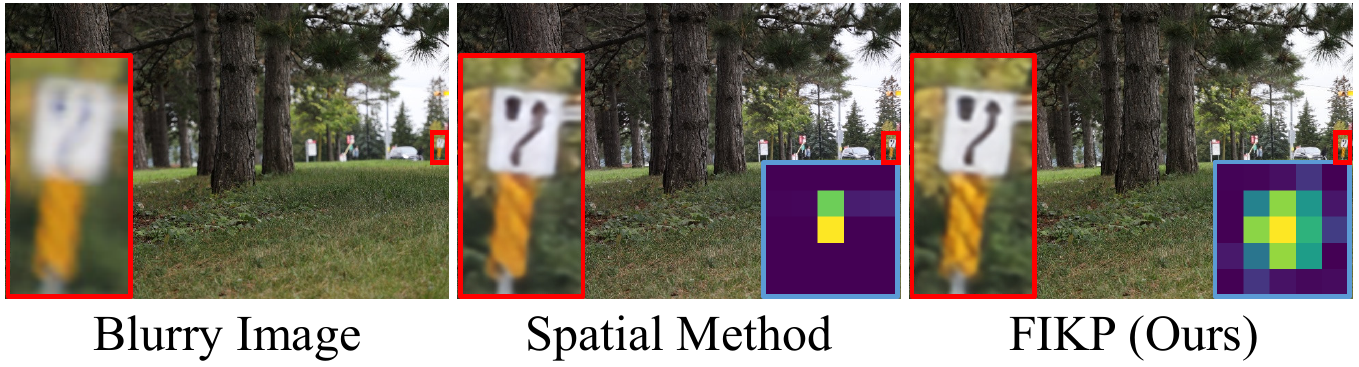}}
\caption{Visualization comparison of deblurring results and predicted kernels between the spatial method and FIKP. Our FIKP better aligns with the kernel prior, reducing artifacts from prediction bias and enhancing deblurring quality.}
\label{fig:kernel_comparison}
\end{figure}

To verify the robustness of frequency approach to missing local high-frequency information, we compare it with the spatial approach.
The spatial method uses Predictor to estimate inverse kernels directly from images, while the frequency method refers to our FIKP.
As shown in the blue box on the lower right of Fig. \ref{fig:kernel_comparison}, the spatial method is easily affected by missing local details, resulting in predicted blur kernels that deviate from the prior assumption \cite{kernel} that defocus kernels approximate a disk shape.
In contrast, our frequency-domain method aligns better with this prior by leveraging global spectral information, leading to more stable and accurate kernel estimation. 
Consequently, as illustrated in the red box on the left of Fig. \ref{fig:kernel_comparison}, it effectively reduces artifacts caused by estimation bias and further improves overall deblurring quality.
Additional ablations on kernel sizes are included in the supplementary.

\subsubsection{Analysis of Dual-Domain Module}
To evaluate the effectiveness of our DDM for spatial-frequency collaborative representation, we design three baseline models for comparison: (a) removing the second stage of DDM to obtain a spatial-only model; (b) removing the first stage to obtain a frequency-only model; and (c) forming a dual-branch structure that processes spatial and frequency features separately with gated fusion.
As shown in Table \ref{tab:DDM}, the spatial-only and frequency-only models achieve PSNR values of 26.24 dB and 26.17 dB, respectively. 
In the dual-branch structure, spatial and frequency features are processed independently, leading to ineffective fusion, resulting in a PSNR of only 26.25 dB. 
In contrast, our DDM adopts a two-stage serial design: the spatial module first restores local details and generates cleaner features for the frequency module, which then focuses on global blur refinement. This sequential strategy avoids feature conflicts and achieves superior deblurring performance, reaching a PSNR of 26.42 dB.
Qualitative comparison results are provided in the supplementary.

\begin{table}
    \centering
    \tabcolsep=0.21cm
    \begin{tabular}{l|c c c c}
    \hline
        Model & PSNR & SSIM  & \# Params (MB) \\
        \hline
        (a) Spatial-only & 26.24 & 0.805 & 3.37 \\
        (b) Frequency-only & 26.17 & 0.804 & 3.73 \\
        (c) Dual-branch & 26.25 & 0.802 & 6.52  \\
        \rowcolor{mygray}
        (d) DDM (Ours) & 26.42 & 0.813 & 5.52 \\
    \hline
    \end{tabular}
    \caption{Results of ablation study on DDM.}
\label{tab:DDM}
\end{table}

\section{Conclusion}
We propose a Frequency-Driven Inverse Kernel Prediction Network (\textbf{FDIKP}) for single image defocus deblurring, leveraging the global modeling strength of the frequency domain.
To improve performance in severely defocused scenarios, we design a Dual-Branch Inverse Kernel Prediction (DIKP) strategy that uses the frequency spectrum as a more discriminative feature space for blur modeling. This alleviates the effect of missing local details and improves inverse kernel estimation.
On top of this, we introduce Position Adaptive Convolution (PAC) to enable pixel-wise deconvolution with a limited set of inverse kernels.
We further develop a Dual-Domain Scale Recurrent Module (DSRM), which jointly models spatial and frequency features to predict coefficient maps and progressively refine results in a scale-recurrent manner.
Extensive experiments demonstrate that our method outperforms existing models on both synthetic and real-world data, especially in restoring fine structural and textural details under large defocus blur.

\section{Acknowledgments}
This work is supported by the CAS Project for Young Scientists in Basic Research (Grant No. YSBR-113).

\bibliography{aaai2026}

\appendix

\section{A: More Implementation and Experiment Details}
\subsection{A1: Datasets}
The benchmark datasets for the SIDD task primarily include five datasets: DPDD \cite{DPDNet}, LFDOF \cite{LFDOF}, RealDOF \cite{IFAN}, RTF \cite{RTF}, and CUHK \cite{CUHK}, with detailed information provided in Table \ref{tab:datasets}. Among them, DPDD and LFDOF can be used for both training and testing, while the remaining datasets are used for testing only. 
It is worth noting that the datasets are not redundantly processed and all models are rigorously trained according to the training data of each dataset.
The LFDOF dataset is not used in the main text experiments; however, the results of the model trained on LFDOF are presented in this supplementary material.

The DPDD dataset \cite{DPDNet} was carefully captured by researchers at York University and the Samsung AI Center, containing 500 scenes. 
It includes 350 scenes in the training set, 74 in the validation set, and 76 in the test set. Each scene consists of one defocus image, two dual-pixel views, and one corresponding all-in-focus image, all with a resolution of 1680$\times$1120.

The LFDOF dataset \cite{LFDOF} is constructed using a light field camera to simulate defocus blur in a physically realistic manner. It contains a total of 11,989 image pairs, including 11,261 pairs for training and 725 pairs for testing.
Each pair of images includes one defocus image and its corresponding all-in-focus image, both with a size of 688$\times$1008.

The RealDOF dataset \cite{IFAN} includes 50 pairs of images captured with a dual-pixel camera system, with each pair of images containing one defocus image and its corresponding all-in-focus image, both with a rough resolution of 2320$\times$1536.

The RTF dataset \cite{RTF} is generated using a light field camera. It consists of 22 pairs of synthesized defocus and sharp images, each with a size of 360×360.

The CUHK dataset \cite{CUHK} contains 704 defocus0 images without corresponding sharp references, and the images vary in resolution. This dataset is constructed from real-world photographs collected from the Internet, covering a diverse range of complex and unconstrained scenes.

\begin{table}[ht]
    \centering
    \begin{tabular}{c|c|c|c|c}
    \hline
        \rotatebox{30}{Datasets} & \rotatebox{30}{\#Images} & \rotatebox{30}{Img. Size} & \rotatebox{30}{Synth.?} & \rotatebox{30}{w/GT?} \\ \hline
         DPDD & 750 & 1680$\times$1120 & \XSolidBrush & \Checkmark \\
        LFDOF & 11989 & 688$\times$1008 & \Checkmark & \Checkmark \\
        RealDOF & 50 & $\approx$2320$\times$1538 & \XSolidBrush & \Checkmark \\
        RTF & 22 & 360$\times$360 & \Checkmark & \Checkmark \\
        \multirow{2}{*}{CUHK} & \multirow{2}{*}{704} & $\approx$640$\times$426 / & \multirow{2}{*}{\XSolidBrush} & \multirow{2}{*}{\XSolidBrush} \\
        & & $\approx$426$\times$640 & &  \\
        \hline
    \end{tabular}
    \caption{Benchmark datasets for single image defocus deblurring task.}
\label{tab:datasets}
\end{table}

\subsection{A2: Loss Function}
Due to space constraints in the main body of the paper, we provide the definitions of the specific loss terms in the single-scale loss function here.

1. \textbf{MSE Loss} (pixel-level loss):
\begin{equation}
    \mathcal{L}_2(y,\hat{y}) = \frac{1}{n} \sum_{i=1}^{n} (y_i - \hat{y}_i)^2,
\end{equation}
where $n$ is the number of samples, $y_i$ and $\hat{y}_i$ are the target and predicted images of the i-th sample, respectively. 

2. \textbf{LPIPS Loss} (perceptual-level loss):
\begin{equation}
    \mathcal{L}_\text{lpips}(y,\hat{y}) = ||\textbf{0}-\text{LPIPS}(y,\hat{y})||_1
\end{equation}
where $\text{LPIPS}(\cdot)$ denotes the perceptual difference between the predicted image $\hat{y}$ and the target image $y$, and $\textbf{0}$ is a zero tensor with the same shape as $\text{LPIPS}(\cdot)$.

3. \textbf{Frequency Loss} (measures the difference in the frequency domain):
\begin{equation}
    \mathcal{L}_\text{freq}(y,\hat{y})=||\mathcal{F}(y)-\mathcal{F}(\hat{y})||_1.
\end{equation}
where, $\mathcal{F}(\cdot)$ represents the 2D fast Fourier Transform.

\begin{table*}
    \centering
    \begin{tabular}{c|c c c | c c c | c c c | c }
    \hline
        \multirow{2}{*}{\textbf{Method}} & \multicolumn{3}{c|}{\textbf{DPDD}} & \multicolumn{3}{c|}{\textbf{RealDOF}} & \multicolumn{3}{c|}{\textbf{RTF}} & \# Params \\
		& PSNR & SSIM & LPIPS & PSNR & SSIM & LPIPS & PSNR & SSIM & LPIPS & (MB)\\ \hline
        Blurry Input & 25.87 & 0.779 & 0.316 & 22.33 & 0.633 & 0.524 & 24.18 & 0.739 & 0.339 & - \\
        DPDNet (2020) & 26.83 & 0.805 & 0.235 & 23.12 & 0.659 & 0.489 & 25.25 & 0.802 & 0.253 & 35.25 \\
        IFAN (2021) & 29.72 & 0.805 & 0.235 & 23.75 & 0.717 & \underline{0.363} & 26.77 & 0.862 & 0.211 & 10.18 \\
        GKMNet (2021) & 29.02 & 0.875 & 0.156 & 23.30 & 0.686 & 0.374 & 26.95 & 0.867 & 0.225 & \textbf{1.41} \\
        DRBNet (2022) & 30.28 & 0.888 & 0.146 & 22.74 & 0.657 & 0.465 & 26.49 & 0.855 & 0.223 & 11.69\\
        Restormer (2022) & 30.49 & 0.891 & 0.139 & 22.35 & 0.634 & 0.519 & 27.12 & 0.877 & 0.181 & 26.10 \\
        NRKNet (2023) & 30.54 & 0.889 & 0.147 & 23.53 & 0.694 & 0.433 & \underline{28.05} & \underline{0.901} & 0.140 & 6.09 \\
        P$^2$IKT (2024) & \underline{30.81} & \underline{0.895} & \underline{0.117} & \underline{23.93} & \underline{0.709} & 0.403 & 27.82 & 0.898 & \underline{0.133} & \underline{3.32} \\
        \rowcolor{mygray}
        \textbf{Ours} & \textbf{31.12} & \textbf{0.899} & \textbf{0.114} & \textbf{24.21} & \textbf{0.736} & \textbf{0.348} & \textbf{28.16} & \textbf{0.909} & \textbf{0.080} & 5.52 \\
        \hline
    \end{tabular}
     \caption{Quantitative comparison results of LFDOF-trained models on \textbf{LFDOF} \cite{DPDNet}, \textbf{RealDOF} \cite{IFAN} and \textbf{RTF} \cite{RTF} dataset. \textbf{Bold} and \underline{underline} indicate the best and second-best results, respectively.}
\label{tab:lfdof_results}
\end{table*}

\subsection{A3: Training Strategy}
We adopt a two-stage training strategy inspired by P$^2$IKT \cite{P2IKT}.
Due to the large gap between the number of images in the DPDD and LFDOF datasets, the number of epochs required for their training is also different, so the specific settings for the two-stage training are also different.

\renewcommand{\dblfloatpagefraction}{.9}
\begin{figure*}[ht]
\centering{\includegraphics[width=\textwidth]{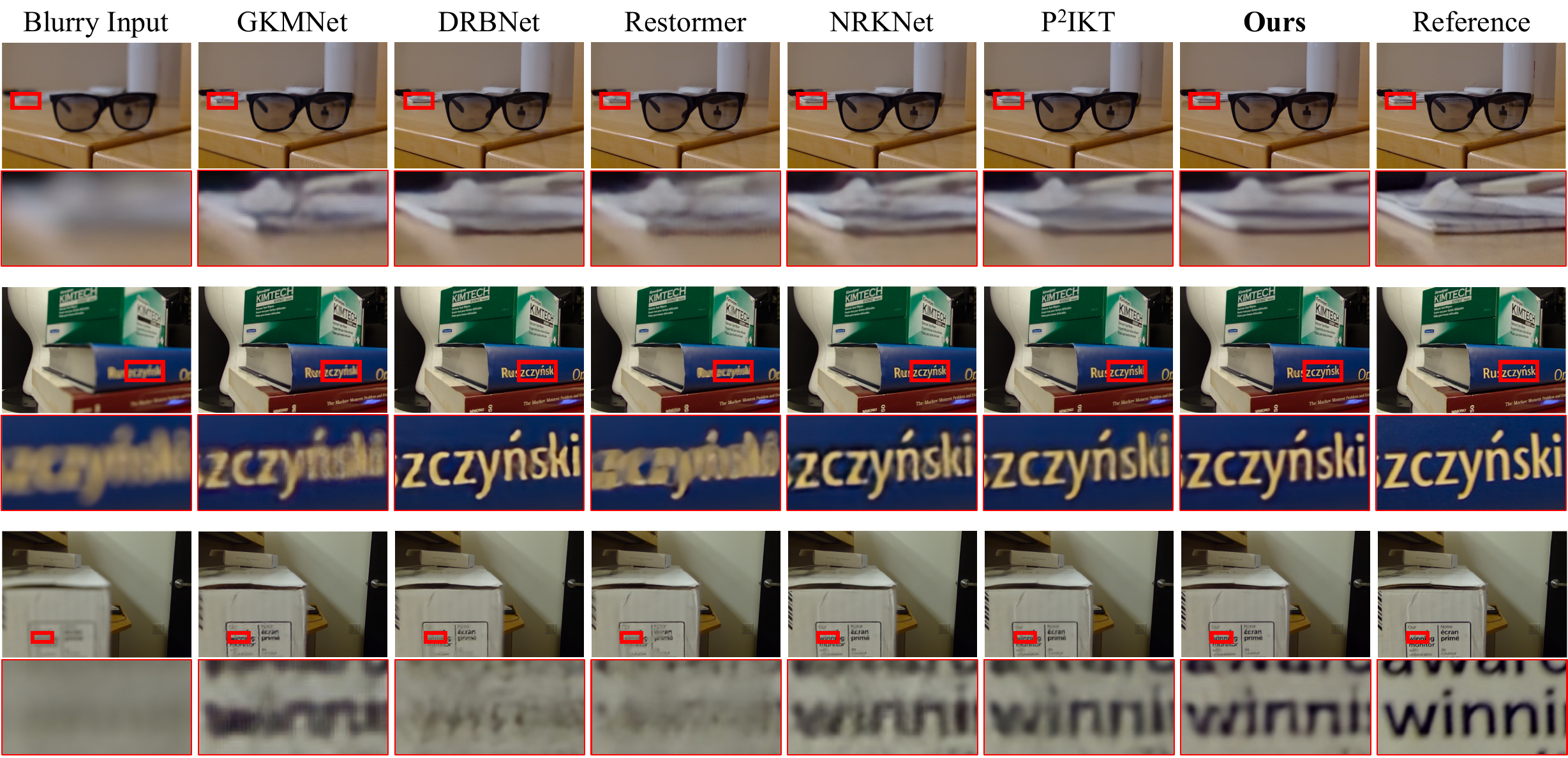}}
\caption{Qualitative results of LFDOF-trained models on \textbf{LFDOF} \cite{LFDOF} dataset.}
\label{fig:lfdof_results}
\end{figure*}

(1) The DPDD-trained model is trained for 1400 epochs with 384$\times$384 inputs and batch size 4 in the first stage, then for 100 epochs with 512$\times$512 inputs and batch size 2 in the second stage.

(2) The LFDOF-trained model is trained for 270 epochs with 256$\times$256 inputs and batch size 4 in the first stage, followed by 30 epochs with 512$\times$512 inputs and batch size 2 in the second stage.

Furthermore, in the second stage, in order to improve the generalization ability of the model, we use a stochastic weight averaging strategy \cite{SWA} to generate the final model for evaluation.

\section{B: Evaluation of LFDOF-trained Model}

\begin{figure*}[ht!]
\centering{\includegraphics[width=\textwidth]{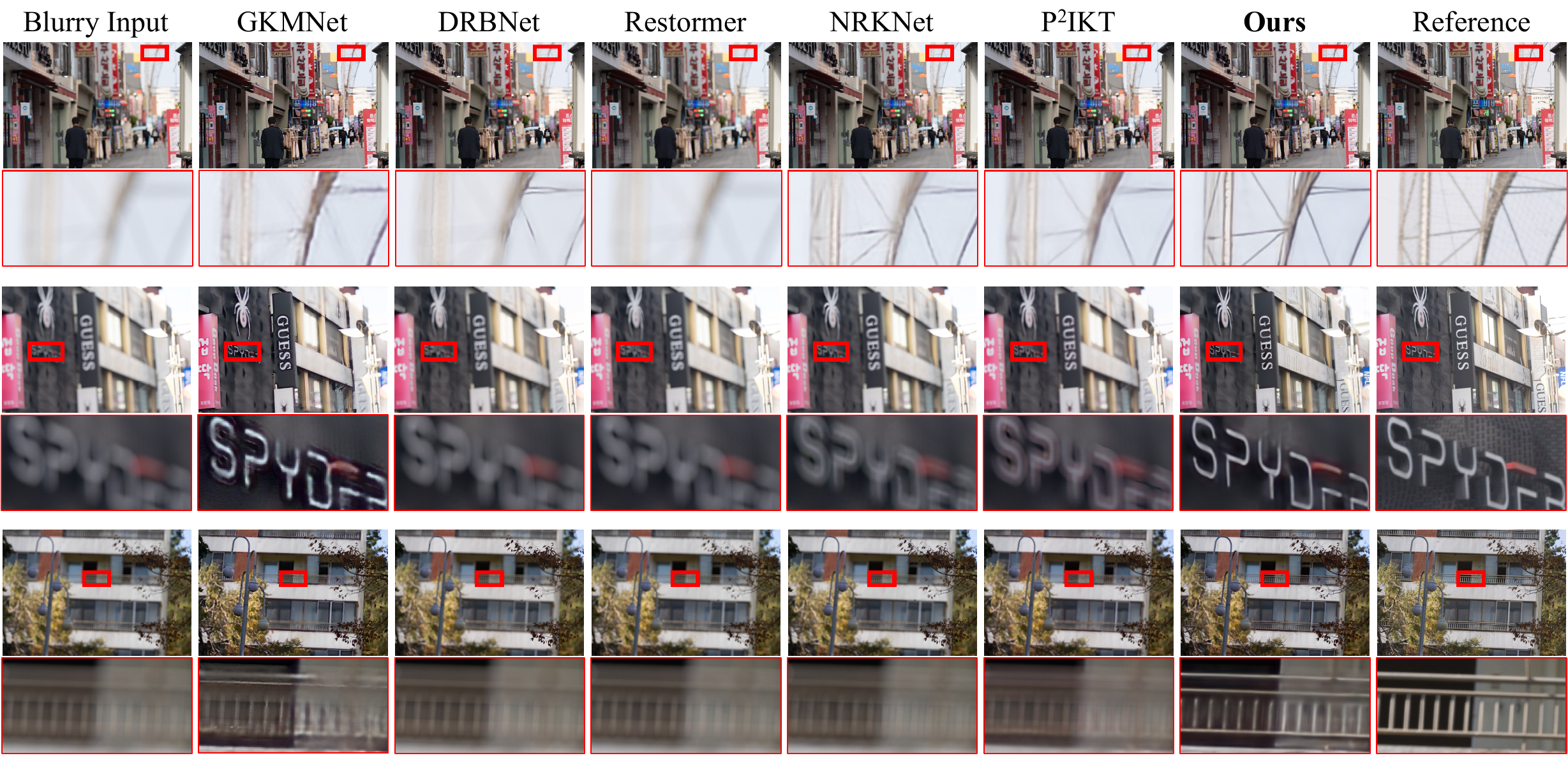}}
\caption{Qualitative results of LFDOF-trained models on \textbf{RealDOF} \cite{IFAN} dataset.}
\label{fig:realdof_results}
\end{figure*}

\renewcommand{\dblfloatpagefraction}{1}
\begin{figure*}[ht!]
\centering{\includegraphics[width=\textwidth]{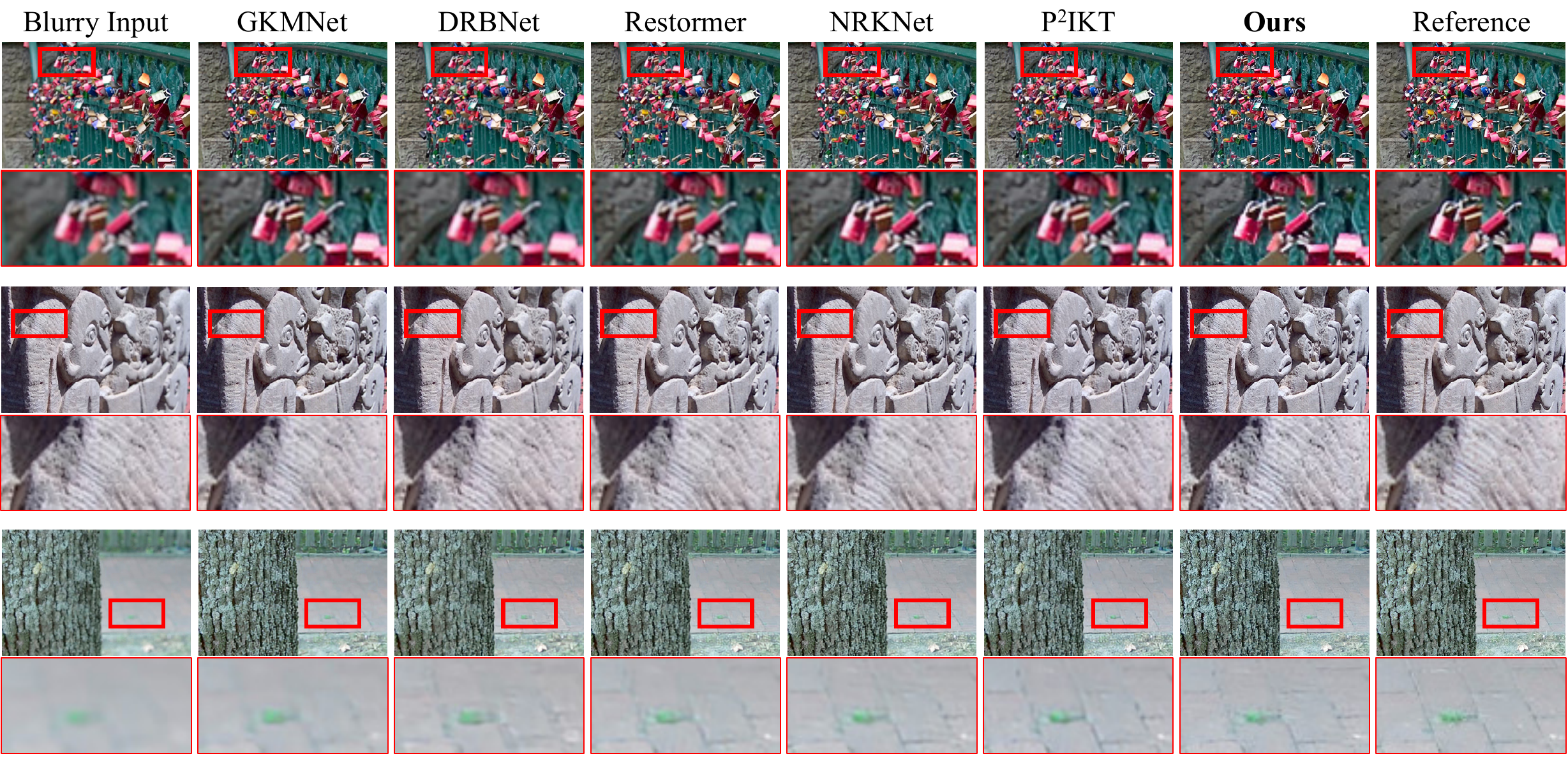}}
\caption{Qualitative results of LFDOF-trained models on \textbf{RTF} \cite{RTF} dataset.}
\label{fig:rtf_results}
\end{figure*}

\renewcommand{\dblfloatpagefraction}{.9}
\begin{figure*}[ht!]
\centering{\includegraphics[width=\textwidth]{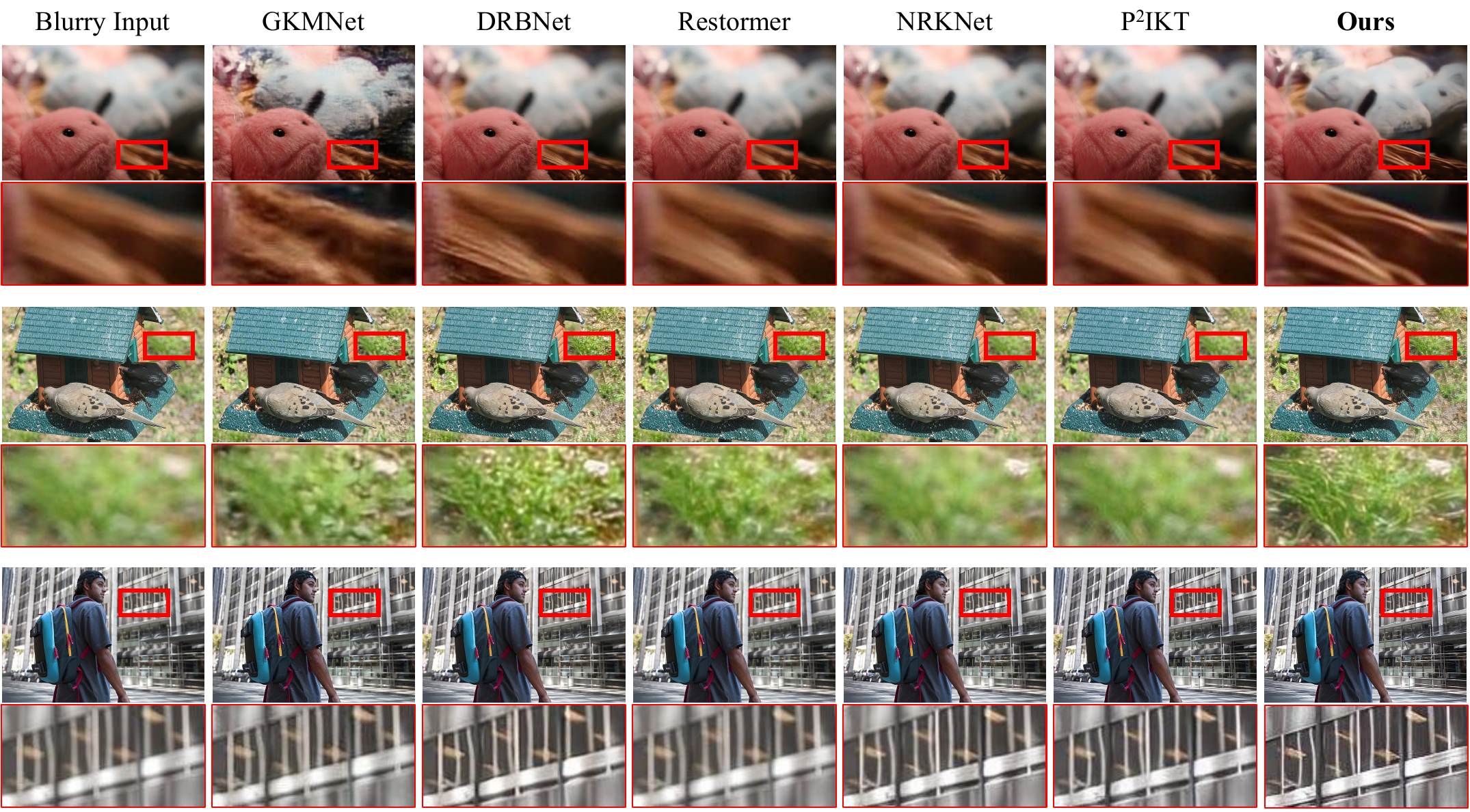}}
\caption{Qualitative results of LFDOF-trained models on \textbf{CUHK} \cite{CUHK} dataset.}
\label{fig:cuhk_results}
\end{figure*}

\begin{table*}
    \centering
    \tabcolsep=0.5cm
    \begin{tabular}{c c c c|c c c c}
    \hline
        $\mathcal{L}_2$ & $\mathcal{L}_\text{lpips}$ & $\mathcal{L}_\text{freq}$ & $\mathcal{L}_\text{MS}$ & PSNR$\uparrow$ & SSIM$\uparrow$ & MAE$\downarrow$& LPIPS$\downarrow$\\ 
        \hline
        \Checkmark & ~ & ~ & ~ & 25.694 & 0.788 & 0.0395 & 0.255\\ 
        \Checkmark & \Checkmark & ~ & ~ & 25.692 & 0.782 & 0.0396& \textbf{0.121}\\
        \Checkmark & ~ & \Checkmark & ~ & 26.258 & 0.808 & 0.0375& 0.217\\
        \Checkmark & \Checkmark & \Checkmark & ~ & 26.329 & 0.811 & 0.0370 & 0.217\\
        \rowcolor{mygray}
        \Checkmark & \Checkmark & \Checkmark & \Checkmark & \textbf{26.420} & \textbf{0.813} & \textbf{0.0366}& 0.185\\
        \hline
    \end{tabular}
    \caption{Results of ablation study about loss function. Bold indicates the best result.}
\label{tab:loss}
\end{table*}

Due to space limitations in the main text, we present here the results of models trained on the LFDOF \cite{LFDOF} dataset.
We compare our FDIKP with several state-of-the-art end-to-end deep learning methods, including DPDNet \cite{DPDNet}, IFAN \cite{IFAN}, GKMNet \cite{GKMNet}, DRBNet \cite{DRBNet}, Restormer \cite{Restormer}, NRKNet \cite{NRKNet}, and P$^2$IKT \cite{P2IKT}. Among these methods, only NRKNet provides pretrained weights on the LFDOF dataset. Therefore, for a fair comparison, we retrained the remaining models on the LFDOF dataset following the same protocol.

The results of the quantitative comparison on LFDOF \cite{LFDOF} dataset are shown in Table \ref{tab:lfdof_results}. 
Our proposed method achieves outperformance of other methods in all three metrics. 
Specifically, it improves 0.31dB, 0.004, and 0.003 over the second-best P$^2$IKT on PSNR, SSIM, and LPIPS, respectively.
The qualitative comparison results are in Fig. \ref{fig:lfdof_results}.
In regions with severe blur, other models often struggle, whereas our method leverages the highly discriminative feature space provided by the frequency domain to accurately estimate the inverse kernel. 
This enables superior deblurring performance and effectively restores structural and textual details.

We also evaluate our model on the RealDOF \cite{IFAN}, RTF \cite{RTF}, and CUHK \cite{CUHK} datasets to verify its generalization capability. The quantitative results on the RealDOF and RTF datasets are shown in Table \ref{tab:lfdof_results}. On the RealDOF dataset, our model achieves the best performance across all metrics, with a PSNR of 24.21 dB, SSIM of 0.736, and LPIPS of 0.348, outperforming the second-best results by 0.28 dB (P$^2$IKT), 0.027 (P$^2$IKT), and 0.015 (IFAN), respectively. On the RTF dataset, our model also achieves the highest PSNR (28.16 dB), SSIM (0.909), and lowest LPIPS (0.080), surpassing the second-best results by 0.11 dB (NRKNet), 0.008 (NRKNet), and 0.053 (P$^2$IKT), respectively.

Fig. \ref{fig:realdof_results} presents the visual comparison results on RealDOF dataset. In regions with severe blur, methods such as DRBNet, Restormer, NRKNet, and P2IKT struggle to restore clear content. Although GKMNet can partially handle such cases, its spatial-domain approach suffers from inaccurate kernel estimation when high-frequency details are missing, resulting in prominent artifacts in the restored images. In contrast, our method leverages frequency-domain features to accurately estimate the inverse kernel, successfully restoring these severely blurred regions without introducing artifacts.
Fig. \ref{fig:rtf_results} shows the visual results on RTF dataset. Even in regions with relatively mild blur, our method outperforms existing approaches, effectively recovering rich texture details in areas such as door locks, walls, and ground surfaces.
Fig. \ref{fig:cuhk_results} shows the visual comparison results on CUHK dataset. For real-world scenes collected from the internet, our method effectively restores image details and textures, such as the woven pattern of the basket, the texture of the grass, and the structural details of the window, without introducing artifacts. This demonstrates the strong generalization ability of our method in complex real-world scenarios.

\section{C: Analysis and Discussion}
Due to the limited space in the main text, we present additional ablation studies in this section to analyze the effect of the loss function, DDM, and kernel size. All ablation experiments are conducted on the DPDD \cite{DPDNet} dataset.

\renewcommand{\dblfloatpagefraction}{.9}
\begin{figure*}[ht]
\centering{\includegraphics[width=\textwidth]{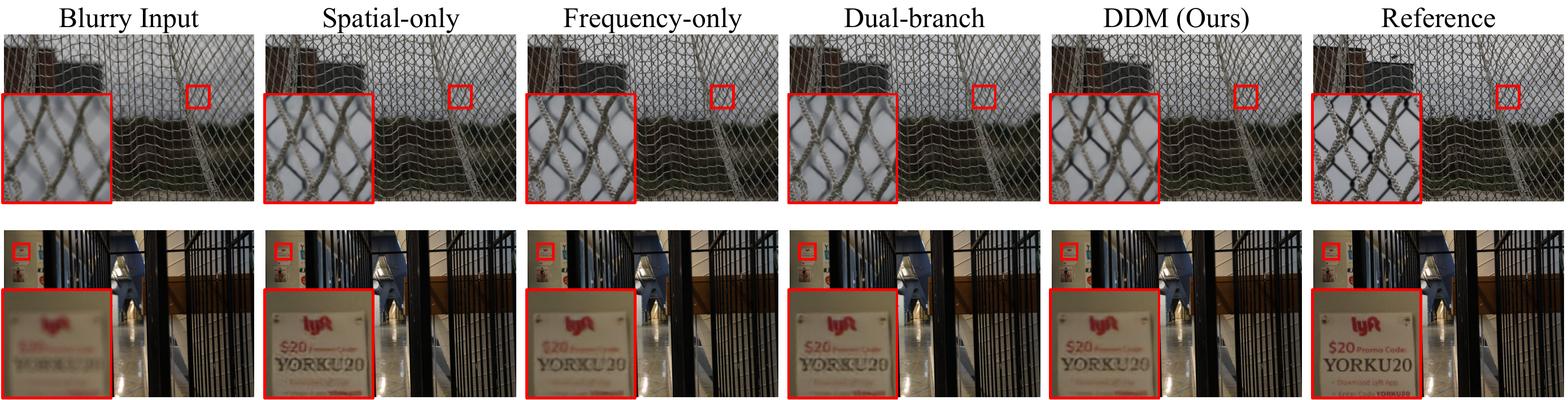}}
\caption{Qualitative comparison results from the ablation study of DDM on \textbf{DPDD} \cite{DPDNet} dataset.}
\label{fig:DDM}
\end{figure*}

\begin{figure}
\centering{\includegraphics[width=0.47\textwidth]{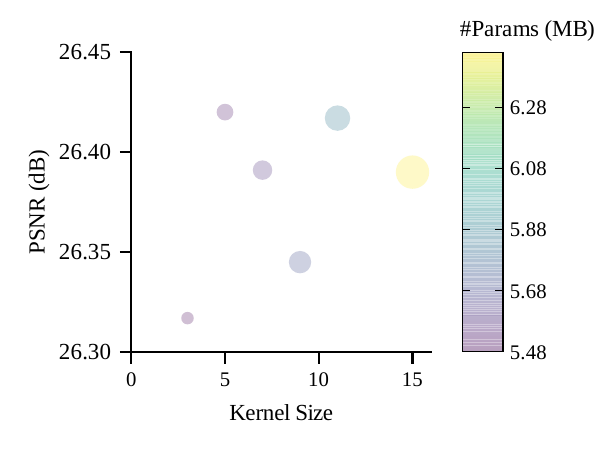}}
\caption{Relationship between kernel size and model performance (PSNR). The size of the scatter indicates the size of the model parameter count as the color bar, which gradually increases as the kernel size increases.}
\label{fig:kernel_size}
\end{figure}

\subsection{C1: Effect of Loss Function}
We deeply analyze the impact of the terms in the single-scale loss function, i.e., the LPIPS loss function and the frequency domain loss function, on the model performance based on the commonly used L2 loss.
As shown in Table \ref{tab:loss}, the introduction of the LPIPS loss function (2nd row) on top of the baseline $\mathcal{L}_2$ leads to a significant improvement in the LPIPS metric, highlighting its effectiveness in aligning high-level semantic features. 
When frequency domain constraints are incorporated alongside the baseline $\mathcal{L}_2$ (3rd row), the model demonstrates notable improvements across all four evaluation metrics, suggesting that these constraints enhance the model's ability to reconstruct high-frequency image details. 
Particularly striking is the joint optimization strategy (4th row), where the combination of LPIPS and frequency domain losses yields a marked synergistic effect, further boosting the model's performance.

Building on this foundation, we incorporate a multi-scale loss strategy to validate its effectiveness. 
As shown in the last row of Table \ref{tab:loss}, adding the multi-scale loss strategy $\mathcal{L}_\text{MS}$ further improves the model's performance on PSNR and SSIM metrics by leveraging the model's multi-output capabilities to optimize from coarse to fine, thereby imposing more direct constraints on the first two stages.

\subsection{C2: Effect of DDM}
We present qualitative comparison results in Fig. \ref{fig:DDM} to further validate the effectiveness of DDM discussed in the main text. 
The results demonstrate that the proposed DDM achieves optimal deblurring performance.
As observed in the second row, purely spatial-domain methods, relying solely on local information, tend to introduce artifacts; frequency-domain methods, while leveraging global information to mitigate artifacts, often produce overly smoothed images with loss of details. 
Dual-branch approaches struggle to fully integrate the advantages of both domains, resulting in limited reconstruction quality. 
In contrast, our method effectively combines spatial and frequency domain information, suppressing artifacts while preserving more image details, thereby achieving a more balanced and higher-quality image restoration.

\subsection{C3: Effect of Kernel Size}
We explore the relationship between predicted kernel size and model performance, where kernel size is set from smallest to largest as 3, 5, 7, 9, 11, and 15, and PSNR is used to measure model performance. 
It is noteworthy that the number of predicted inverse kernels equals the kernel size and changes accordingly.
As shown in Fig. \ref{fig:kernel_size}, the number of parameters of the model increases as the predicted kernel size becomes larger (indicated by the size of the scatter in the figure), but the PSNR value does not have a clear linear relationship with the kernel size.

When the kernel is small (e.g., kernel size = 3), it is difficult for the model to capture and recover the low-frequency components (i.e., the overall structure and smooth regions) in the image due to the small receptive field, which leads to blurring of the image. 
In contrast, when the blurring kernel is large (e.g., kernel size $\geq$ 7), the model's receptive field increases and is able to capture a larger range of structural information, but at the same time, it will tend to smooth the image and suppress the high-frequency components (e.g., edges and textures), which leads to loss of image details.

At a kernel size of 5, the model reaches a maximum value of 26.420 dB in PSNR, indicating that at this point the model achieves an optimal balance between capturing global information and preserving details. 
At the same time, this kernel size corresponds to the second smallest number of model parameters, indicating that a good trade-off between performance and complexity is also achieved. 
Therefore, the final choice for this paper is to set the kernel size to 5.

\begin{figure}[ht]
\centering{\includegraphics[width=0.47\textwidth]{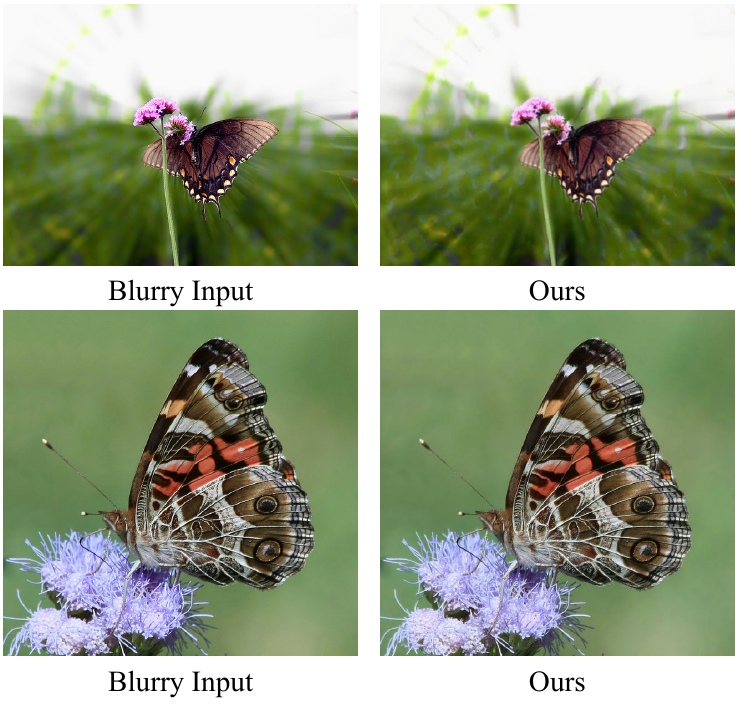}}
\caption{Failure cases.}
\label{fig:failure_cases}
\end{figure}

\section{D: Visualizations}

\subsection{D1: Failure Cases}
Our proposed FDIKP leverages the global nature of frequency domain information to learn the defocus kernel and model global structural details. 
However, it struggles in certain cases. 
For instance, in the case of radial blur (shown in the first row of Fig. \ref{fig:failure_cases}), the defocus kernel properties are not well-defined, which limits our model's performance. 
Similarly, when the blur is so severe that the information is almost completely overwhelmed, such as the background blur in the second row of Fig. \ref{fig:failure_cases}, where the image appears almost entirely green, our model cannot effectively restore the image, even with the frequency domain's global characteristics.

\subsection{D2: More Deblurring Results}
\begin{figure*}
\centering{\includegraphics[width=\textwidth]{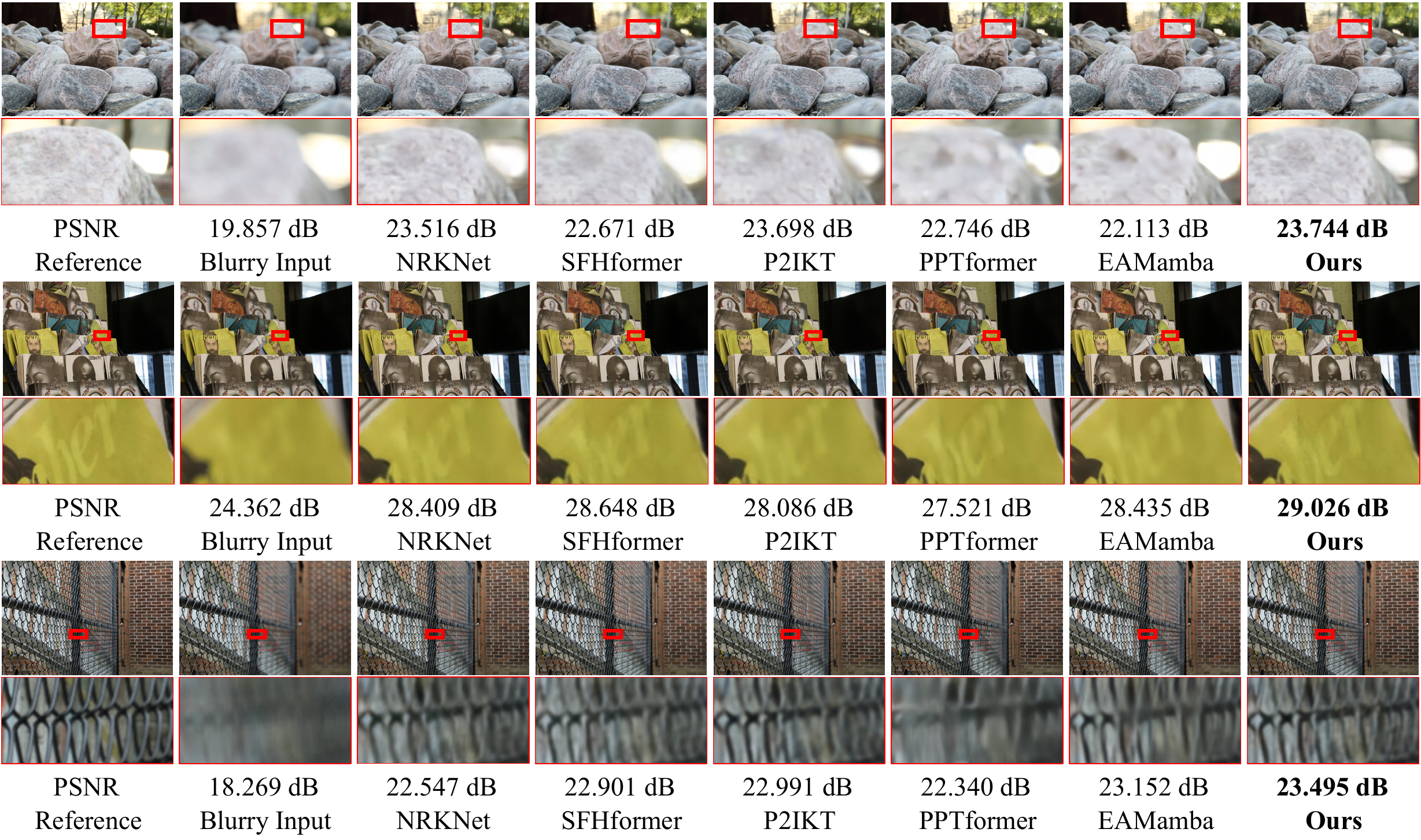}}
\caption{Qualitative results of DPDD-trained models on \textbf{DPDD} \cite{DPDNet} dataset.}
\label{fig:dpdd_results}
\end{figure*}

\begin{figure*}
\centering{\includegraphics[width=\textwidth]{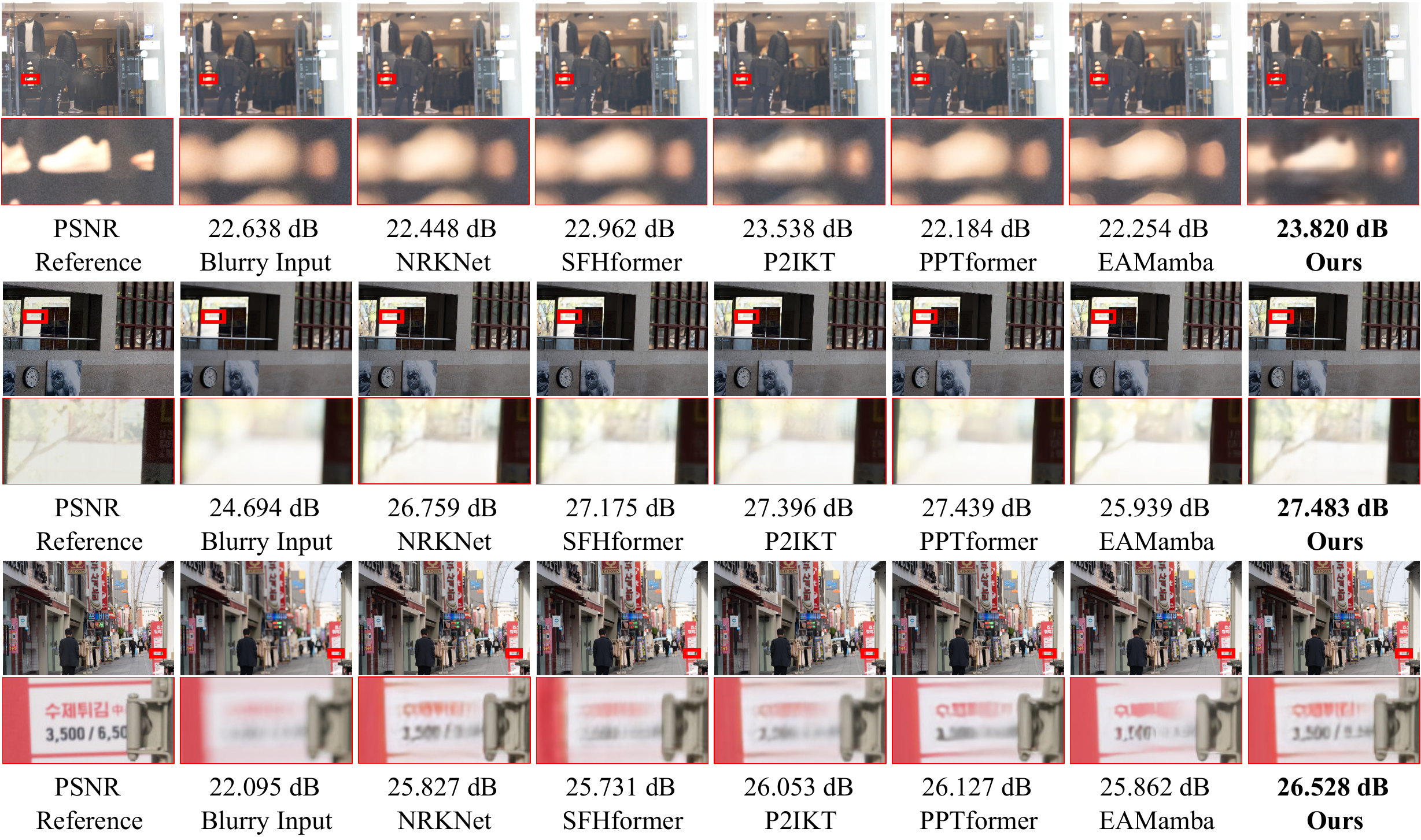}}
\caption{Qualitative results of DPDD-trained models on \textbf{RealDOF} \cite{IFAN} dataset.}
\label{fig:realdof2_results}
\end{figure*}

\begin{figure*}
\centering{\includegraphics[width=\textwidth]{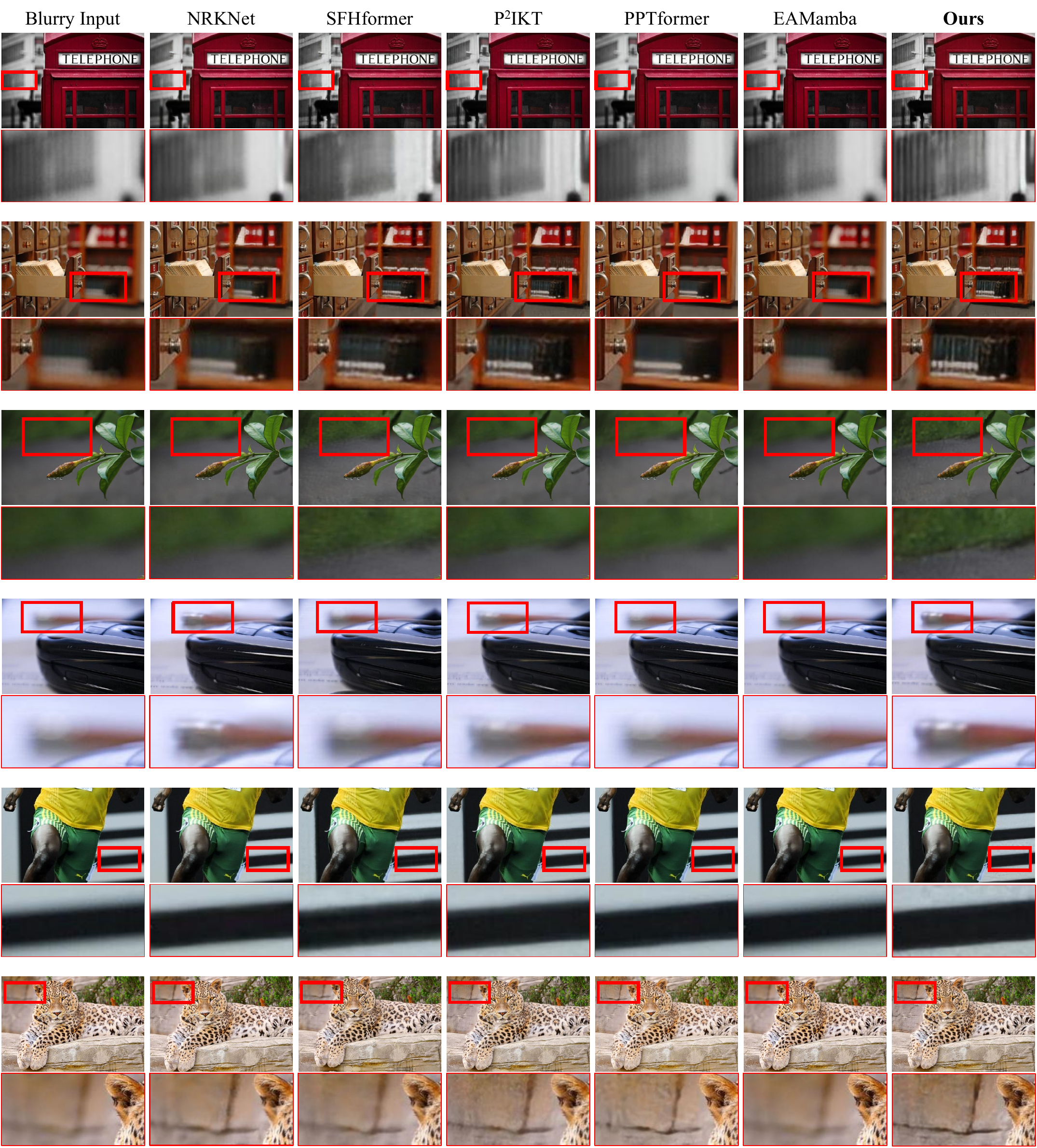}}
\caption{Qualitative results of DPDD-trained models on \textbf{CUHK} \cite{CUHK} dataset.}
\label{fig:cuhk2_results}
\end{figure*}

Due to space limitations in the main text, additional visual comparisons of the DPDD-trained models, including NRKNet \cite{NRKNet}, SFHformer \cite{SFHformer}, P2IKT \cite{P2IKT}, PPTformer \cite{PPTFormer}, EAMamba \cite{EAMamba}, and our method, on the DPDD \cite{DPDNet}, RealDOF \cite{IFAN}, and CUHK \cite{CUHK} datasets are presented in Fig. \ref{fig:dpdd_results}, \ref{fig:realdof2_results}, and \ref{fig:cuhk2_results}, respectively.

\renewcommand{\dblfloatpagefraction}{.9}

\renewcommand{\dblfloatpagefraction}{.9}

\end{document}